\documentclass[lettersize,journal]{IEEEtran}
\usepackage{amsmath,amsfonts}
\usepackage{algorithmic}
\usepackage{algorithm}
\usepackage{array}
\usepackage[caption=false,font=normalsize,labelfont=sf,textfont=sf]{subfig}
\usepackage{textcomp}
\usepackage{stfloats}
\usepackage{url}
\usepackage{verbatim}
\usepackage{graphicx}
\usepackage{cite}

\usepackage{booktabs}
\usepackage{multirow}

\begin{document}

\title{Edge-Enabled Real-time Railway Track Segmentation}

\author{Chenglin Chen, Fei Wang, Min Yang, Yong Qin, Yun Bai 
        % <-this % stops a space
%\thanks{This paper was produced by the IEEE Publication Technology Group. They are in Piscataway, NJ.}% <-this % stops a space
%\thanks{Manuscript received April 19, 2021; revised August 16, 2021.This research was supported by 1) the State Key Laboratory of Rail Traffic Control and Safety under Grant No.RCS2023K007, Beijing Jiaotong University, and 2) Guangzhou Municipal Science and Technology Project under Grant 2023A03J0011. The first and second corresponding authors are Yun Bai and Yong Qin respectively.}
%\thanks{Chenglin Chen, Fei Wang, and Min Yang are with the Thrust of Intelligent Transportation, The Hong Kong University of Science and Technology (Guangzhou), Guangzhou 511400, China (e-mail: cchen363@connect.hkust-gz.edu.cn, fwang423@connect.hkust-gz.edu.cn,  myang945@connect.hkust-gz.edu.cn ).}
%\thanks{Yong Qin is with the State Key Laboratory of Rail Traffic Control and Safety, Beijing Jiaotong University, Beijing 100044, China (e-mail: yqin@bjtu.edu.cn).}
%\thanks{Yun Bai is with the Thrust of Intelligent Transportation and Guangdong Provincial Key Lab of Integrated Communication, Sensing and Computation for Ubiquitous Internet of Things, The Hong Kong University of Science and Technology (Guangzhou), Guangzhou 511400, China;  the State Key Laboratory of Rail Traffic Control and Safety, Beijing Jiaotong University, Beijing 100044,China (e-mail: yunbai@hkust-gz.edu.cn).}
}

\markboth{Journal of \LaTeX\ Class Files,~Vol.~14, No.~8, August~2023}%
{Shell \MakeLowercase{\textit{et al.}}: A Sample Article Using IEEEtran.cls for IEEE Journals}

\IEEEpubid{0000--0000/00\$00.00~\copyright~2021 IEEE}

\maketitle

\begin{abstract}
Accurate and rapid railway track segmentation can assist automatic train driving and is a key step in early warning to fixed or moving obstacles on the railway track. However, certain existing algorithms tailored for track segmentation often struggle to meet the requirements  of real-time and efficiency on resource-constrained edge devices. Considering this challenge, we propose an edge-enabled real-time railway track segmentation algorithm, which is optimized to be suitable for edge applications by optimizing the network structure and quantizing the model after training. Initially, Ghost convolution is introduced to reduce the complexity of the backbone, thereby achieving the extraction of key information of the interested region at a lower cost. To further reduce the model complexity and calculation, a new lightweight detection head is proposed to achieve the best balance between accuracy and efficiency. Subsequently, we introduce quantization techniques to map the model’s floating-point weights and activation values into lower bit-width fixed-point representations, reducing computational demands and memory footprint, ultimately accelerating the model’s inference. Finally, we draw inspiration from GPU parallel programming principles to expedite the pre-processing and post-processing stages of the algorithm by doing parallel processing. The approach is evaluated with public and challenging dataset RailSem19 and tested on Jetson Nano. Experimental results demonstrate that our enhanced algorithm achieves an accuracy level of 83.3\% while achieving a real-time inference rate of 25 frames per second when the input size is 480×480, thereby effectively meeting the requirements for real-time and high-efficiency operation.
\end{abstract}


\section{Introduction}
\IEEEPARstart{I}{n} the contemporary urban transportation system, rail transit assumes an indispensable role as an efficient and reliable mode of transportation, providing a solid foundation for the sustainable development of cities and the smooth flow of traffic. An efficient environmental perception system is a pivotal factor in ensuring the safe operation of trains, enhancing transportation efficiency, and mitigating accident risks. Specific environmental perception tasks encompass the detection of track conditions, identification of obstacles, and recognition of railway signaling components such as signal lights, switches, and signal cables\cite{phusakulkajorn2023artificial,liu2019review,tang2022literature}. Among them, the railway track segmentation algorithm bears paramount significance within the context of the railway perception system. The primary objective of this task is to proficiently extract the structural elements of railway tracks from intricate visual scenes, rendering them independent within the visual representation from their surroundings. The successful execution of this task not only contributes to the prevention of collisions between trains and potential obstacles, thereby averting accidents and potential harm to individuals, but it also furnishes critical support for train operations, track maintenance, and intelligent scheduling. 

The railway track segmentation algorithm must exhibit prompt responsiveness and reliability for environmental perception to ensure the safety and efficacy of the railway transportation system. Given the substantial volume of sensitive data associated with railway transportation, including train locations, operational statuses, and monitoring images, the deployment of railway track segmentation algorithms on edge devices becomes imperative. This approach allows for localized data processing, thereby reducing data transmission latency and the risks of data leakage, thus enhancing data privacy and security. \IEEEpubidadjcol Furthermore, edge devices typically require fewer hardware resources, consequently reducing hardware costs. Nevertheless, it is worth noting that edge devices often contend with constrained computational and storage resources, posing a challenge for existing railway track segmentation algorithms to meet real-time requisites. Thus, the imperative task at hand is the development of railway track segmentation algorithms capable of meeting the real-time decision-making and response requirements upon edge devices. In light of above background, we propose a lightweight track segmentation algorithm founded on YOLOv8-seg and integrating advanced technologies such as pruning and quantization to implement an efficient and real-time track segmentation algorithm on edge devices. Concisely, the contributions of this work are delineated follows.
\begin{enumerate}
\item{Constructed a lightweight network tailored for railway scenarios based on YOLOv8-seg. To the best of our knowledge, in the railway domain, this approach is the first track segmentation algorithm that can achieve real-time performance on resource-constrained edge devices.}
\item{Incorporated Ghost module to optimize feature extraction network. Additionally, we have proposed a light decoupled head, to further reduce model complexity, and achieve an optimal balance between precision and efficiency.}
\item{Quantized the trained model, converting floating-point weights and activation values into fixed-point representations, to boost inference speed. Furthermore, inspired by GPU parallel programming principle, CUDA parallel algorithm is used to accelerate pre-processing and post-processing.}
\item{Conduced comprehensive experimental evaluation on optimized YOLOv8-seg, demonstrating the algorithm performance through various criteria.}
\end{enumerate}

The remainder of this paper is structured as follows: Section II introduces some related works. Section III presents the proposed methodology. Section IV provides the experiment results. Finally, a conclusion is presented in Section V.

\section{RELATED WORKS}
Currently, designing a real-time railway track segmentation system for rail transit remains a challenging task. On one hand, existing research generally tends to design complex models to meet high accuracy requirements in railway scenarios. However, this trend toward high accuracy often comes at the expense of speed. For safety considerations in the field of rail transit, in addition to high accuracy, low latency is also a mandatory criterion that algorithms must meet. This means that the algorithm must respond quickly to ensure the safety of train operations. On the other hand, edge devices equipped on trains often have limited computing capabilities. Therefore, research efforts towards the deployment of lightweight models with high accuracy, fast inference, and low latency suited for edge devices become crucial and pressingly needed to ensure the feasibility of the system in rail transit.

\textbf{Railway Track Segmentation.} The railway track segmentation algorithm is extensively employed across a range of domains, including railway maintenance, automated train operation, and track monitoring. In the early development stages of railway track segmentation, numerous scholars dedicated their research efforts  using traditional computer vision methods. For instance, some researchers employed techniques such as Canny, Hough transform, and mathematical morphology for the detection and segmentation of railway tracks \cite{singh2019vision,karakose2017new}. Some others combined the use of Gabor wavelets at different scales to enhance the identification of  rail edges in conjunction with noise filtering \cite{selver2016camera}. The extraction of left and right tracks can also be achieved by applying dynamic programming and Hough transformation \cite{nassu2011rail}. Nassu et al. \cite{kang2023tpe} carried out an approach that performs rail extraction by matching edge features to candidate rail patterns modeled as sequences of parabola segments. All these methods are simple to calculate but are often sensitive to illumination changes and struggling to deal with complex scenes. In addition, they heavily rely on handcrafted features and may not generalize well to diverse real-world scenarios. 

 In recent years, with the advancements in deep learning techniques and computing powers, traditional vision methods are gradually being displaced by learning-based methods. Owing to the superior feature learning capabilities inherent to convolutional neutral network (CNN), learning-based methods have the capability to automatically acquire effective features and adapt to environmental changes, resulting in higher accuracy and robustness in railway track segmentation tasks. In the case of deep learning, image segmentation comprises two critical subtasks: semantic segmentation and instance segmentation. Semantic segmentation aims to predict the semantic label of each pixel in an image. Prominent algorithms in this direction include FCN \cite{dai2016r}, DeepLab \cite{chen2017deeplab}, U-Net \cite{ronneberger2015u}, SegNet \cite{badrinarayanan2017segnet}, and BiSeNet \cite{yu2018bisenet}. Instance segmentation can be seen as a combination of object detection and semantic segmentation, where the model needs to not only produce a pixel-wise segmentation map of the image, but also assign a specific object instance for each pixel. Noteworthy algorithms for instance segmentation includeMask R-CNN \cite{he2017mask}, YOLACT \cite{bolya2019yolact}, and SOLO \cite{wang2020solo}. Among all algorithms, the YOLO \cite{terven2023comprehensive} series of algorithms has consistently gained prominence due to its outstanding real-time performance and ease of engineering application. The latest YOLOv8\cite{jocher2023yolo} algorithm has also expanded the YOLOv8-seg component to support segmentation tasks. Compared to other segmentation algorithms, this algorithm demonstrates superior performance in terms of real-time processing and suitability for deployment. However, it still falls short of satisfying the real-time demands imposed by low-performance edge devices.
 
 The majority of existing railway track segmentation works are derived from these general algorithms, optimized for and tailored to the characteristics of rail  transportation. Some researchers \cite{kang2023tpe,wang2019railnet,li2020vanishing, li2020railnet} directly employ neural networks for the segmentation of railway track areas. Certain scholars \cite{yang2021topology} have explored methods for detecting paired rails by leveraging the geometric features of railways, although this approach exhibits limited robustness, as minor changes within the track area can lead to misjudgments of the track. Although deep learning has made significant advancements in the field of general segmentation algorithms, there still exist several limitations when it comes to the railway track segmentation One of the primary reasons for this is the closed nature of the railway industry. Due to railway data security and privacy concerns, publicly available rail datasets are relatively rare, and the data quality is not high, which limits the training and generalization of deep learning models on railway track segmentation tasks. Another challenge is that most of algorithms have high complexity, making it difficult to meet real-time requirements, let alone deploying them on edge devices.
 
 \begin{figure*}[!t]
 	\centering
 	\includegraphics[width=7in]{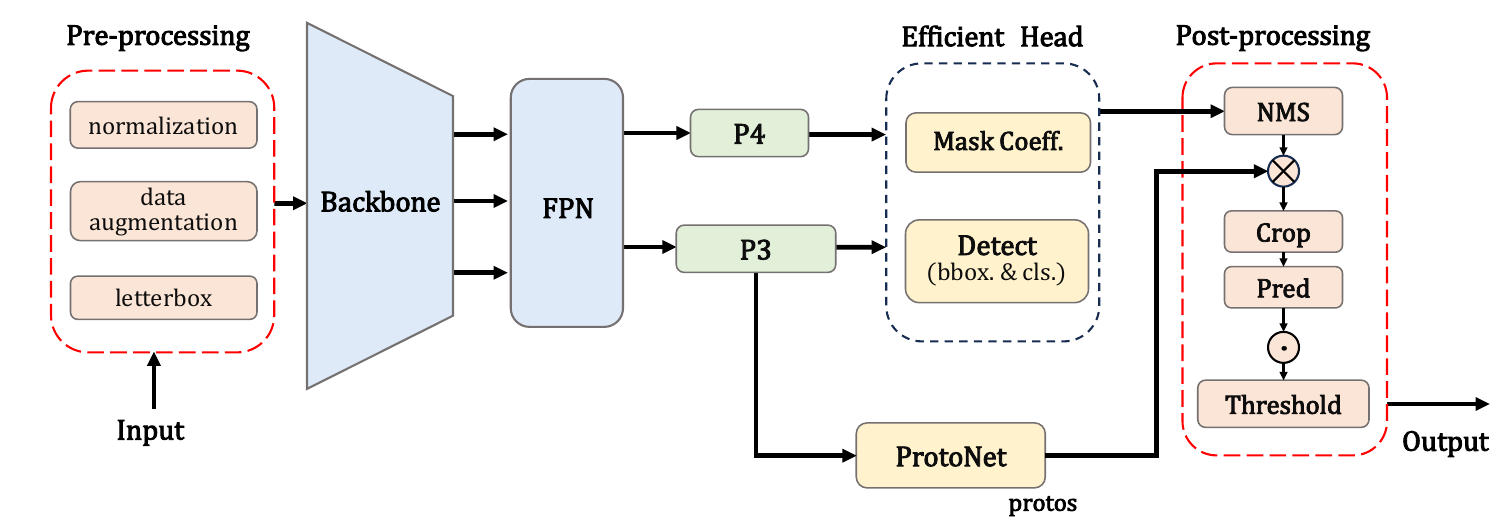}
 	\caption{The pipeline of proposed method. {P3, P4} are denoted as the feature maps with different scale generated by FPN.}
 	\label{fig_1}
 \end{figure*}
 
  \begin{figure*}[!t]
 	\centering
 	\includegraphics[width=7in]{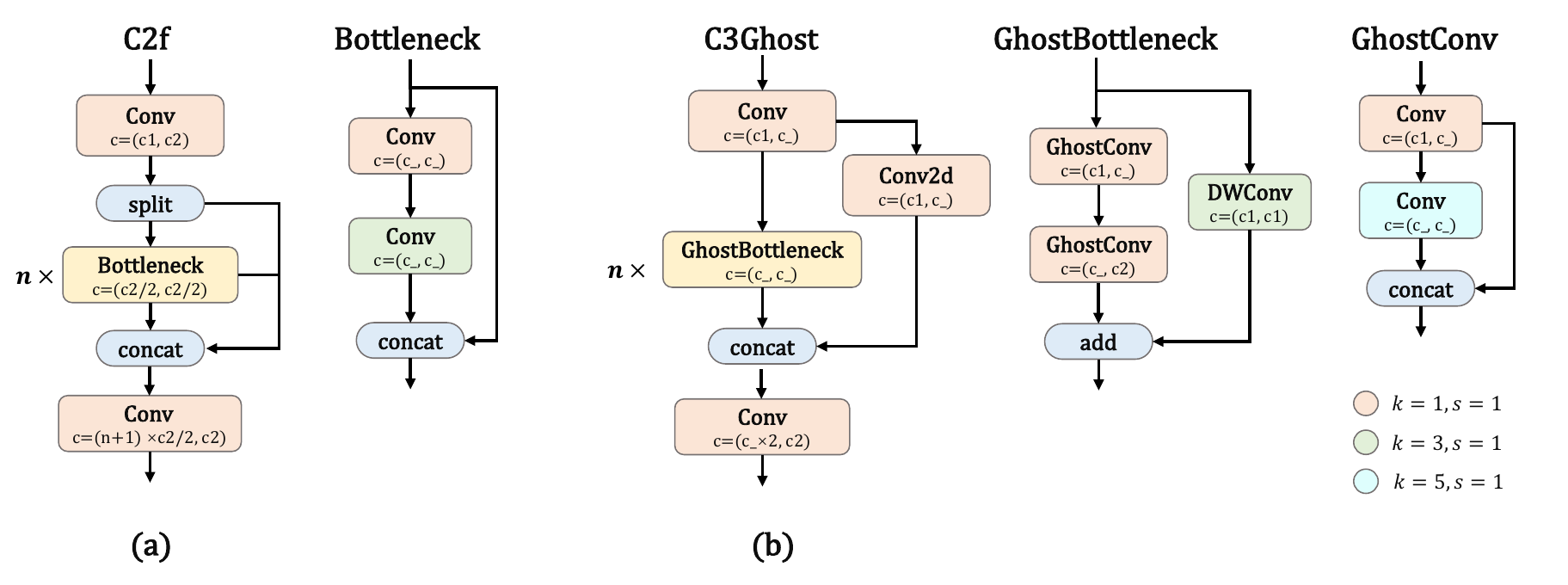}
 	\caption{The network structure of G2f with Bottleneck, compared with C3Ghost with GhostBottleneck.}
 	\label{fig_2}
 \end{figure*}

\textbf{ Model Acceleration and Compression.} Model compression and acceleration are quite broad and active research areas in deep learning, aiming to speed up inference and reduce the computation, energy, and storage costs to adapt to resource-constrained environments such as embedded devices, mobile applications, and real-time systems.  So far, common solutions for model acceleration and compression include network pruning \cite{liu2017learning, xu2020convolutional, liang2021pruning}, network quantization \cite{liang2021pruning,gholami2022survey,wang2023computation}, low-rank decomposition \cite{denton2014exploiting,jaderberg2014speeding}, knowledge distillation \cite{chen2017learning,yang2022focal}, and lightweight model design \cite{sandler2018mobilenetv2,iandola2016squeezenet,ma2018shufflenet,han2020ghostnet,mehta2018espnet}. Among them, pruning, quantization, and lightweight model design dominate in engineering practices. Lightweight network architectures are designed to reduce model parameters and complexity while maintaining model performance at certain levels. The design focuses on efficient convolution kernels and backbone networks. For example, SqueezeNet \cite{iandola2016squeezenet} designs a compact model that primarily uses 1×1 convolutions instead of 3×3 convolutions to reduce the parameters. MobileNet \cite{sandler2018mobilenetv2} employs depth-wise separable convolutions instead of standard convolutions and uses a width multiplier for more efficient use of model parameters. ShuffleNet \cite{ma2018shufflenet} introduces the concepts of group convolution and channel shuffle to reduce computational complexity by rearranging channels. ESPNet \cite{mehta2018espnet} focuses on lightweight semantic segmentation tasks and utilizes a method called "Lightweight Branch Selection" to minimize calculations and parameters. GhostNet \cite{han2020ghostnet} proposes a novel Ghost module to generate more feature maps from cheap operations. This module can be taken as a plug-and-play component to build lightweight network easily. Additionally, NASNet \cite{zoph2018learning} and EfficientNet \cite{tan2019efficientnet} perform model optimization through neural architecture search. Among them, they all made great efforts to strike a balance between model accuracy and efficiency.
 
Apart from the design of lightweight network architectures, pruning and quantization are two other major branches of model compression and acceleration. Pruning is a technique for removing redundant parameters or neurons from a neural network with a minimal impact on accuracy. Note that pruning is typically performed during or after training, and a strategy is required to determine which parts of the network can be pruned. Quantization is the process of converting the model parameters from high-precision representations (e.g., 32-bit floating-point, FP32, the size usually used to store parameters) to lower-precision (e.g., 16-bit floating-point, FP16 or 8-bit integer, INT8) representations,  as resource-constrained edge devices are limited in computing such models with high-precision parameters. In summary, pruning and quantization are general techniques for optimizing deep learning models. They are designed to address resource constraints and real-time needs and can be applied to various application domains, although they sometimes result in reduced accuracy. But as far as we know from reviewing the literature, there are very few algorithms that apply above technologies to fully accelerate railway track segmentation. Therefore, in this paper, we integrate the aforementioned techniques to enhance the performance of railway track segmentation. We introduce Ghost modules and modify the model head to construct a lightweight network. However, given that pruning entails the need for retraining and hand-crafted parameters, we prefer to use quantization to speed up model inference rather than pruning. Additionally, we leverage parallel processing to accelerate both the pre- and post-processing stages of the model.  Our approach achieved an exceptional performance  on low-performance edge devices.
 
\section{METHODOLOGY}
In this section, we delve into the detailed development of the proposed lightweight model and various strategies and techniques for model acceleration.
\subsection{System Overview}
We proposed a lightweight instance segmentation model based on YOLOv8-seg, which utilizes the YOLACT. Fig. \ref{fig_1} gives an overview of the proposed approach. This approach consists of two stages, namely lightweight instance segmentation and model acceleration. Initially, the model required for training is developed through an instance segmentation network. Subsequently, the trained model is quantized to achieve inference acceleration. Simultaneously, the GPU parallelism strategy is applied to accelerate both model pre-processing and post-processing. 

For the model training stage, the input images are fed into the light backbone and FPN to obtain two diverse sized feature maps P3 and P4, which then are fed into the detection and mask coefficients branches respectively. The detection branch outputs categories (cls.) and bounding boxes (bbox.), while the mask coefficients branch generates k (default k is 32) mask coefficients ranging from -1 to 1. Only the lowest-level features P3 enter the ProtoNet branch, which is used to generate k prototype masks. We perform Non-Maximum Suppression (NMS) to suppress the predictions from both the mask coefficients and detection branches. Afterward, we multiply the remaining predictions by the masks and sum them to obtain the instance segmentation results. This method achieves a balance between accuracy and speed of railway track segmentation.
\subsection{Encoder}
The core components of the backbone and neck in YOLOv8-seg are the C2f modules. The specific structure of the C2f module is illustrated in Fig. \ref{fig_2}(a), which consists primarily of 3x3 convolutions and 1x1 convolutions. However, in deep convolutional networks, stacking numerous convolutional layers consumes a substantial number of parameters and computational resources. Additionally, the C2f module employs many skip connections and split operations to enhance connections between different convolutional layers. While these operations facilitate information flow, they also increase convolutional complexity. Furthermore, operations like "split" are not particularly hardware friendly. To effectively reduce model complexity, we replaced many C2f modules with C3Ghost modules, as depicted in Fig. \ref{fig_2}(b). 
\begin{figure}[!t]
	\centering
	\includegraphics[width=3.5in]{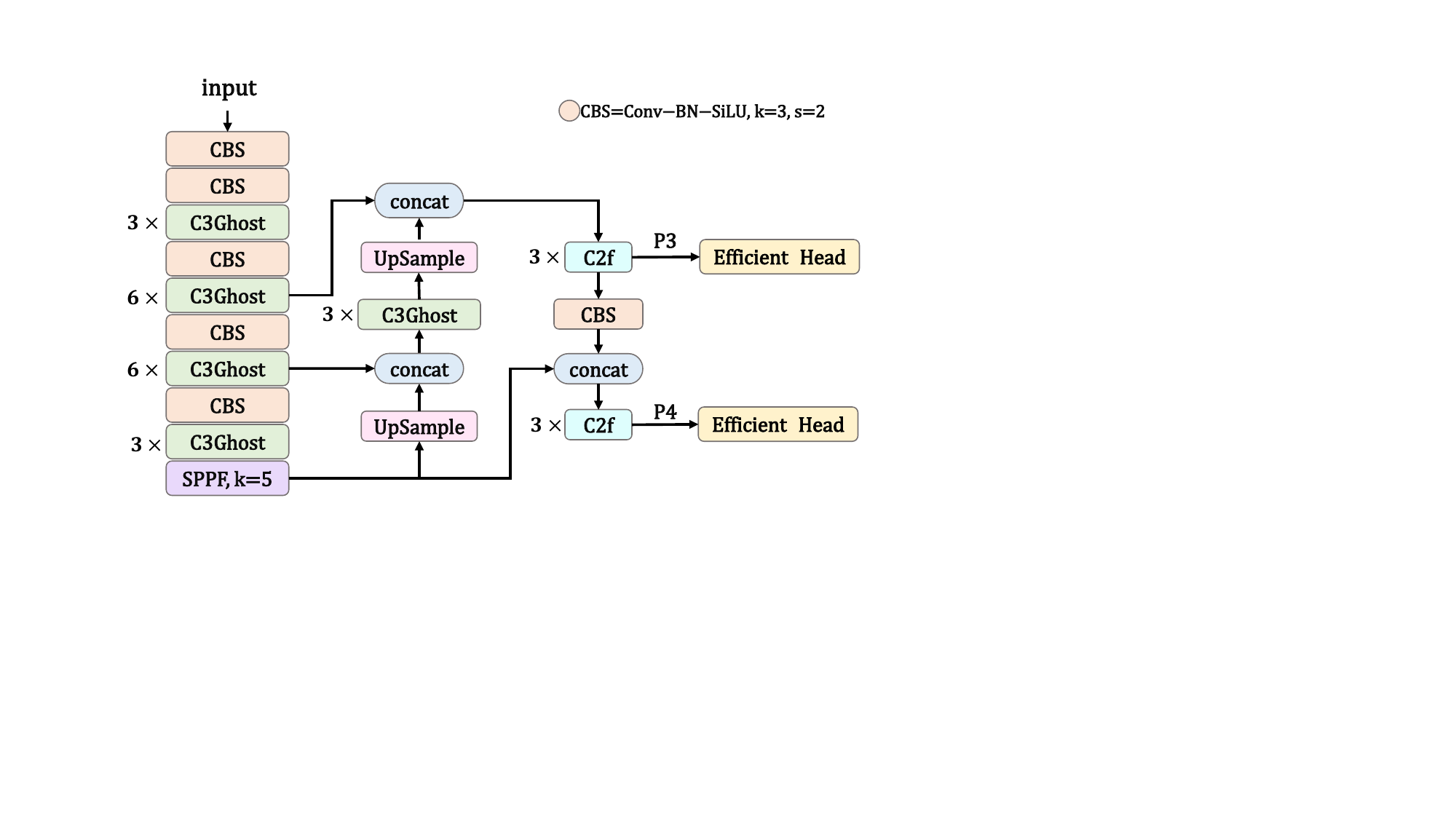}
	\caption{Details of the encoder structure (backbone and neck).}
	\label{fig_3}
\end{figure}
\begin{figure}[!t]
	\centering
	\includegraphics[width=3in]{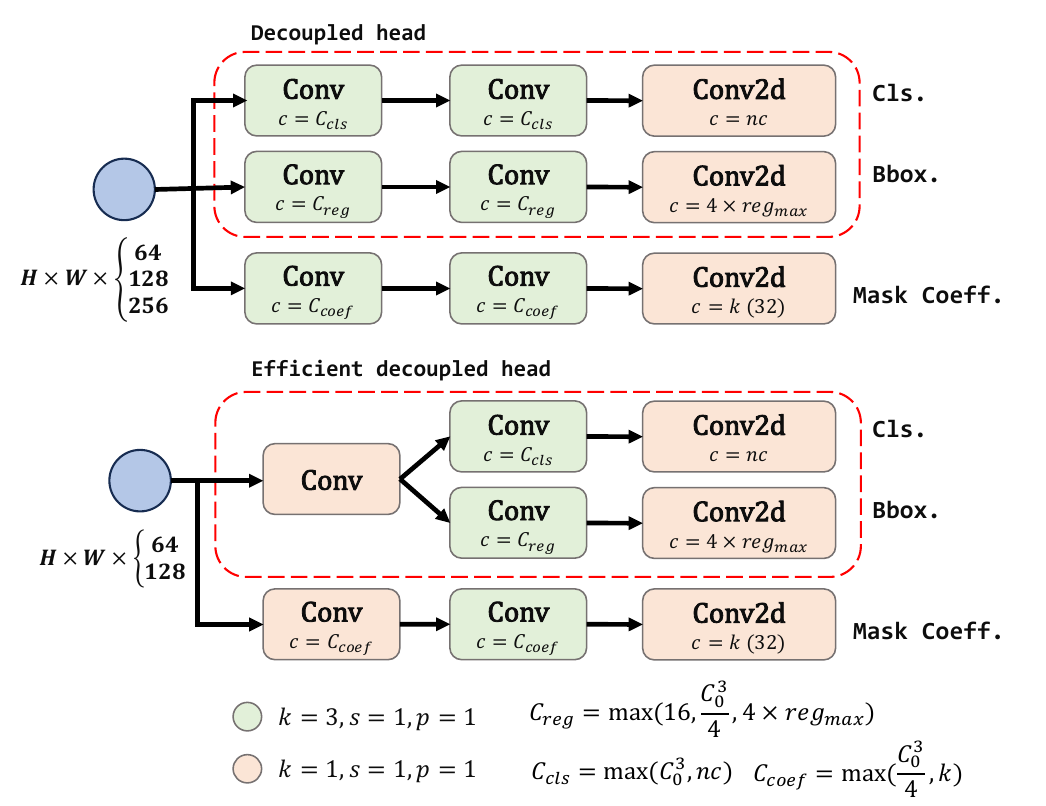}
	\caption{Details of original decoupled head (top) and light weighted decoupled head (down).}
	\label{fig_4}
\end{figure}
\begin{figure}[!t]
	\centering
	\includegraphics[width=2.5in]{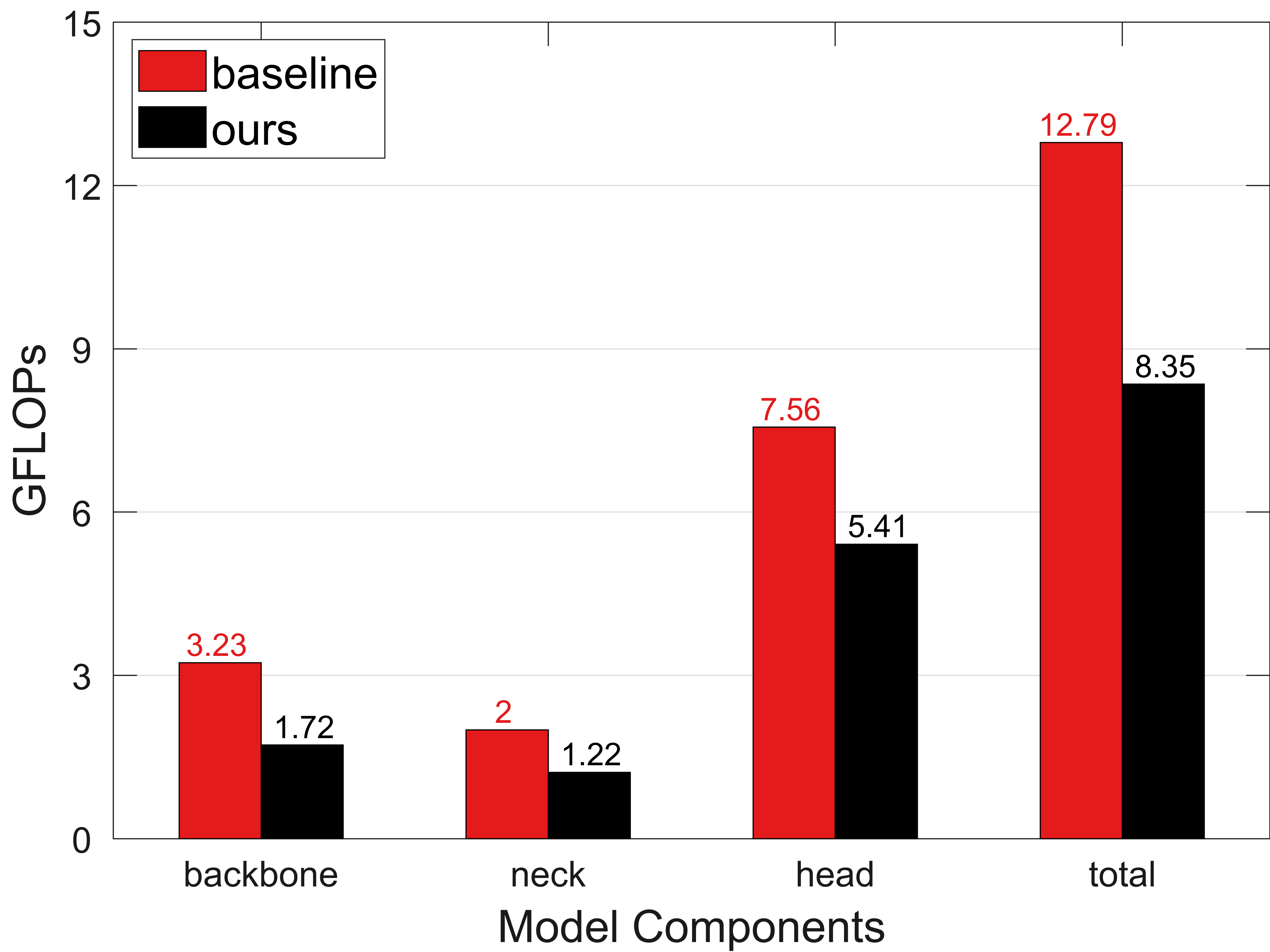}
	\caption{Details of original decoupled head (top) and light weighted decoupled head (down).}
	\label{fig_mc}
\end{figure}
The C3Ghost module is similar to the C3 module used in YOLOv5 but replaces all bottlenecks with GhostBottleneck components. C3Ghost reduces skip connections and eliminates the use of split operations. GhostBottleneck is primarily constructed using grouped convolutions and Ghost modules. The core idea behind this component is to reduce the demand for computational resources by combining a small number of convolutional kernels with more cost-effective linear transformation operations (such as grouped convolutions), without compromising model performance. In summary, the optimized encoder can learn features with reduced number of trainable parameters, thereby accelerating computation. The specific improved encoder structure is illustrated in Fig. \ref{fig_3}.
\subsection{Light Head}
YOLOv8-seg employs a decoupled head to extract object position and category information separately. Considering the distinct concern of classification and regression, network performance can be improved by learning regression and classification through different branches. However, this approach introduces some extra inference costs compared to a coupled head. By analyzing the computational load of each network layer, as shown in Fig. \ref{fig_mc}, it can be found that GFLOPs at the head dominate. The bulk of the computational load in this part comes primarily from the 3×3 convolutions within the decoupled head, which results in a significant increase in inference cost as the number of channels and input size increases. In order to reduce computational load, we propose a more lightweight decoupled head with fewer channels and convolutional layers, as depicted in Fig. \ref{fig_4}. In detection branch, we first reduce the feature channels to 64 through a 1×1 convolution and then feed them into separate decoupled classification and regression branches. The number of convolutional layers in both branches is reduced from 2 to 1. For the mask coefficients branch, similar to the classification branch of the detection head, we replace the original two 3×3 convolutions with a combination of 1x1 and 3x3 convolutions, and then enter the classification head to predict mask scores. This optimization strategy not only resolves the conflict of inconsistent emphasis between classification and positioning, but also effectively reduces the number of parameters and computational complexity of head.

In order to detect object at different scales, YOLOv8-seg uses three feature layers of different scales (P3, P4, P5) for prediction. Typically, lower-level features are used to detect small objects, while higher-level features are used for larger objects. However, in our application scenario, data is often obtained from a forward-facing camera fixed on a train, and the object scale changes within a small range. To further reduce the model complexity, we abandon P5 and retain P3 and P4 feature layers. The final experiment proved that this strategy sacrificed a negligible amount of accuracy in exchange for a huge reduction in the number of model parameters and computations. The comparison of the computation of model components before and after optimization is shown in Section IV. C.

\subsection{Quantization}
We speed up the trained model further by employing post-training quantization (PTQ). PTQ represents a lightweight quantization technique that reduces model size without compromising accuracy as much as possible. Notably, PTQ entails low engineering effort and computational cost, does not require retraining, and uses little or no data for hyperparameter tuning. Optimization strategies mainly include layer \& tensor fusion, precision calibration, dynamic tensor memory and so on. After obtaining the trained model, we primarily adopt the first two optimization strategies, that is, merging convolution-batch normalization-activation, and eliminating Concat layers. These operations effectively reduce the number of operators, making the model smaller and faster. Subsequently, we turn our attention to quantize the tensor precision of the model. During the model training process, tensors within the network are maintained at high-precision FP32, primarily due to the demands of back propagation. However, during inference, where the absence of back propagation renders high precision unnecessary, judicious reductions in data precision become feasible. Using lower precision requires less memory and enables faster computation. In this study, we use a mixture of symmetric and asymmetric quantization to quantize the precision of weights and activations into FP16. We do not enable dynamic shapes as the image size captured by the forward-facing camera in our scenario remains constant. Therefore, static shapes perfectly align with our requirements and offer faster inference speed. We utilize the TensorRT to implement model optimization operations.
\subsection{Acceleration of pre- and post-processing}
A typical AI model deployment process comprises three stages: pre-processing, inference, and post-processing. Generally, model inference is typically executed on a GPU or dedicated hardware (e.g., NVIDIA Jetson, AIxBoard) by utilizing inference frameworks like TensorRT, OpenVINO. Conversely, the pre-processing and post-processing stages are carried out on a CPU. For computer vision tasks, pre-processing and post-processing operations often require substantial CPU resources and are time-consuming. This is particularly evident on embedded platforms. Consequently, migrating these operations to the GPU can significantly enhance the overall execution efficiency of the entire workflow. 

During the image pre-processing stage, our focus lies in accelerating the normalization and letterbox operations. Normalization involves dividing the pixel values of the image by 255 to obtain floating-point data. Since the original image size typically differs from the required input size of the model, image size transformation is needed to fit into the model input. Letterbox is a technique that resizes the image while maintaining its aspect ratio, padding the margin with 0 pixels. In this paper, we employ affine transformations (translation and scale) along with bi-linear interpolation to achieve a similar resizing effect, preserving the aspect ratio while filling the margin with a fixed value. This approach effectively implements the functionality of letterbox.

After performing model inference on the input data, we obtain the bounding boxes, mask coefficients, and prototype masks. Post-processing is necessary to obtain the desired mask. In the post-processing stage, NMS is primarily employed to remove redundant detection bounding boxes. Subsequently, the desired mask is obtained by conducting element-wise multiplication between the mask coefficients and the prototype mask. Furthermore, the desired  mask is resized to match the dimensions of the original image. These operations are efficiently implemented using parallel coding to accelerate the computational process.

To fully leverage GPU acceleration throughout the entire pipeline, we not only apply the TensorRT to quantize model on the GPU, but also perform the pre- and post-processing stages on the GPU. This ensures end-to-end GPU acceleration for the entire workflow of the proposed method.

\section{EXPERIMENTS}
\subsection{Implementation Details}
\textbf{Datasets.} RailSem19 is an open-source dataset for semantic rail scene understanding, taken from the ego-perspective of trains and trams. It provides extensive annotations in different formats, including geometry-based (rail-relevant polygons, all rails as polylines) and dense label maps. Our model exclusively focuses on the railway track area within the images. Therefore, we customized RailSem19 to obtain the desired data, resulting in 8,000 images. These were subsequently divided into training and validation sets with a 7:3 ratio.

\textbf{Environment Setup.} We conducted experiments on  GeForce RTX 3060Ti with 8GB of GDDR6 memory and a 12th Gen Intel(R) Core(TM) i7-12700F@2.1GHz CPU for training and validation. To further verify the performance of our approach for edge computing devices with lower computing power, we deployed and tested our approach on Jetson Nano, which has a 128-core Maxwell GPU, a 4-core ARM A57@1.43GHz CPU,  and 4GB of LPDDR4 of memory. The compute capability of the RTX3060 Ti and Jetson Nano are 8.6 and 5.3, respectively.

Our approach is based on publicly available PyTorch implementations of YOLOv8-seg and uses development tools such as TensorRT 8.6, and CUDA to optimize the overall model architecture. During the inference stage, we code the process in both Python and C++. Note that the computer with RTX 3060Ti uses TensorRT 8.6, CUDA 11.7, and Jetson nano uses jetpack4.6, with built-in TensorRT 8.2, CUDA 10.2.

\textbf{Training and Inference Setup.} During training, images are cropped to 640×640. We use the pretrained weights of the YOLOv8-seg trained on the COCO-seg \cite{lin2015microsoft}. We use random flip and random clipping to augment the datasets. The SGD optimizer is applied for optimization. The initial learning rate is 0.01, the momentum parameter is 0.9, the weight penalty is 0.0001. Training epochs is set to 100 and batch size is 8.  During inference, no data augmentation is applied. The intersection over union (IoU) threshold of non-maximum suppression (NMS) is set to 0.25, and the confidence threshold is set to 0.05.

\textbf{Evaluation Metrics.} We employ the mean Average Precision (mAP) under 0.5 IoU thresholds as the evaluation metrics on accuracy, as well as Params, GFLOPs, cost and FPS as the ones on computational overhead. Additionally, in resource-constrained scenarios such as embedded devices or edge computing, the following performance metrics also need to be considered, namely throughput, latency, and model size. These indicators reflect the efficiency and deploy ability of the model. In neural networks, the meaning of these indicators is as follows.
\begin{itemize}
\item{Throughput: the maximum number of input samples that a model can process per unit time, typically measured in query per second (qps), responding to images per second.}
\item{Latency: the time spent executing the model, conventionally measured in milliseconds (ms).}
\item{Model Size: the space required for model parameter storage, including weights, bias, and other trainable variables, usually expressed in Megabyte (MB).}
\end{itemize}
Latency refers to the time taken for the model to infer a single image, excluding the time required for pre- and post-processing, whereas the cost metric includes them. FPS denotes the inference frames the model can perform in a second on average, calculated by dividing one second by the cost. In addition, it should be emphasized that the relationship between Latency and the FPS of model inference is not a simple inverse correlation, as their computation often involves different threads. Generally, the execution of Latency is single-threaded, as it measures the inference time of a single image. In contrast, the computation of FPS is typically multi-threaded, as it considers the scenario of processing multiple images within a second. 

\renewcommand\arraystretch{1.3}
%经典三线表
\begin{table*}[!t]
	\caption{Performance Comparison of baseline and proposed method on Railsem19 using GTX3060 Ti}\label{tab:c1} 
	\centering
	\begin{tabular}{ccccccc}
		\toprule
		& $mAP^{Det}_{50}$ & $mAP^{Mask}_{50}$ & FPS (bs=1/8) & GFLOPs & Params & Model Size (MB)\\
		\midrule 
		YOLOv8n-seg & 0.92 & 0.847 & 177.3/370.4 & 12.8	&3.41M	&6.8 \\ 
		Our proposed model  & 0.911 & 0.833 & 196.9/416.7&8.0	&1.39M	&3.0\\  
		\bottomrule
	\end{tabular}
\end{table*}

\begin{table*}[]
	\caption{Performance comparison of model under different device, precision, and input size}\label{tab:c2} 
	\centering
	\begin{tabular}{ccccccc}
		\toprule
		device & precision & input size & throughput (qps) & latency (ms) & cost (ms) & FPS    \\
		\toprule
		\multirow{6}{*}{\begin{tabular}[c]{@{}c@{}}{\centering}GTX 3060 Ti\\ SM=86\end{tabular}} & \multirow{3}{*}{FP32} & 640×640    & 524.2            & 2.3          & 2.50      & 400.0  \\
		&                       & 480×480    & 762.0            & 1.5          & 1.79      & 558.7  \\
		&                       & 224×224    & 1300.3           & 0.8          & 1.26      & 793.7  \\
		
		\cline{2-7}
		& \multirow{3}{*}{FP16} & 640×640    & 957.2            & 1.4          & 1.52      & 657.9  \\
		&                       & 480×480    & 1343.8           & 1.0          & 1.24      & 806.5  \\
		&                       & 224×224    & 1964.9           & 0.6          & 1.00      & 1000.0 \\
		
		\hline
		\multirow{6}{*}{\begin{tabular}[c]{@{}c@{}}Jetson Nano\\ SM=53\end{tabular}} & \multirow{3}{*}{FP32} & 640×640    & 14.6             & 68.4         & 73.70     & 13.6   \\
		&                       & 480×480    & 24.8             & 40.3         & 46.08     & 21.7   \\
		&                       & 224×224    & 96.1             & 10.4         & 24.35     & 41.1   \\
		
		\cline{2-7}
		& \multirow{3}{*}{FP16} & 640×640    & 18.1             & 55.1         & 57.32     & 17.4   \\
		&                       & 480×480    & 30.3             & 32.9         & 39.95     & 25.0   \\
		&                       & 224×224    & 115.8            & 8.6          & 21.27     & 47.0   \\
		\toprule
	\end{tabular}
\end{table*}

\subsection{Comparisons}
We evaluated the accuracy and efficiency of our approach by comparing it with the baseline yolov8n-seg on RailSem19. As shown in Table \ref*{tab:c1}, the proposed method achieves mAPs of 0.911 and 0.833 on detection and instance segmentation tasks, respectively, which are only reduced by 0.009 and 0.014 compared to the baseline, while offering comparable or superior performance in terms of FPS, model size, GFLOPs and Params. Model size has been reduced from 6.8M to 3.0M, saving 55.9\% of storage space. In addition, the baseline model inference speed is promoted around 12\%, while saving 37.5\% GFLOPs, indicating its effectiveness in accelerating model. The inference speed differs significantly when the batch size is set to 1 and 8. This is because when the batch size equals 8, we utilize the dataset class function from PyTorch, which enables easy implementation of multi-threaded data prefetching and batch preloading, thus speeding up the overall execution time. However, in real-world edge development scene, depending on the task load, the number of thread resources actually allocated for our tasks may be just one. Hence, we need to consider the model inference speed when the batch size is set to 1, that is, using single-threaded data iteration. This is a major contributor to the performance disparity. Unless specifically stated in the following sections, all results are calculated with a batch size of 1 representing typical edge computing scenario.

Our model demonstrates competitive speed across various platforms, as illustrated in Table \ref{tab:c2}. This can be attributed to our effective network design, alongside a sophisticated approach to quantization and GPU parallelism. In Table \ref{tab:c2}, we present a comparative analysis of the performance of our proposed model under varying conditions, including device type, quantization precision, and input size. We have observed an inverse relationship between throughput and input image size, as well as a direct relationship between latency and image size. Specifically, as input image size decreases, there is a corresponding increase in throughput and a decrease in latency. There are several factors, including system load, that can introduce minor variations in different runs. Consequently, even when tested on the same device, the FPS obtained from the same model may exhibit slight discrepancies. In order to mitigate the impact of these minor fluctuations, we conducted an extensive series of tests, with 1000 runs on the GTX 3060Ti and 10 runs on the Jetson Nano. By calculating the average result from these tests, we were able to ensure the reliability of our outcomes.

\textbf{Performance analysis of the impact of input size on Jetson Nano.} As shown in Table \ref{tab:c2}, we evaluated the performance of our model with FP32-precision on the Jetson Nano under different input size cases, specifically 640×640, 480×480, and 224×224. The model achieved FPS of 17.4, 25, and 47 respectively. Considering the low-performance nature of the Jetson Nano, the inference speed of our model can be deemed relatively fast.

\textbf{Analysis of the speedup effects on different devices.} As can be seen from Table \ref{tab:c2} as well, the inference speed of our model (excluding pre- and post-processing time) using FP16-precision on the Jetson Nano is approximately 1.2 times that of FP32, whereas on the GTX 3060Ti, it is 1.6 times faster.  These speedup ratios fall short of the theoretical doubling, suggesting that the acceleration effects of the same model can differ across devices with varying computational power. Generally, devices with greater computational power exhibit more pronounced acceleration. This differential acceleration effect is influenced by numerous factors, including the versions of CUDA, CUDNN, PyTorch, and TensorRT. Therefore, to achieve optimal performance, it is crucial to consider these factors in a holistic manner when selecting hardware and software configurations.

\begin{figure*}[!t]
	\centering
	\subfloat[Original Image]{\includegraphics[width=2in]{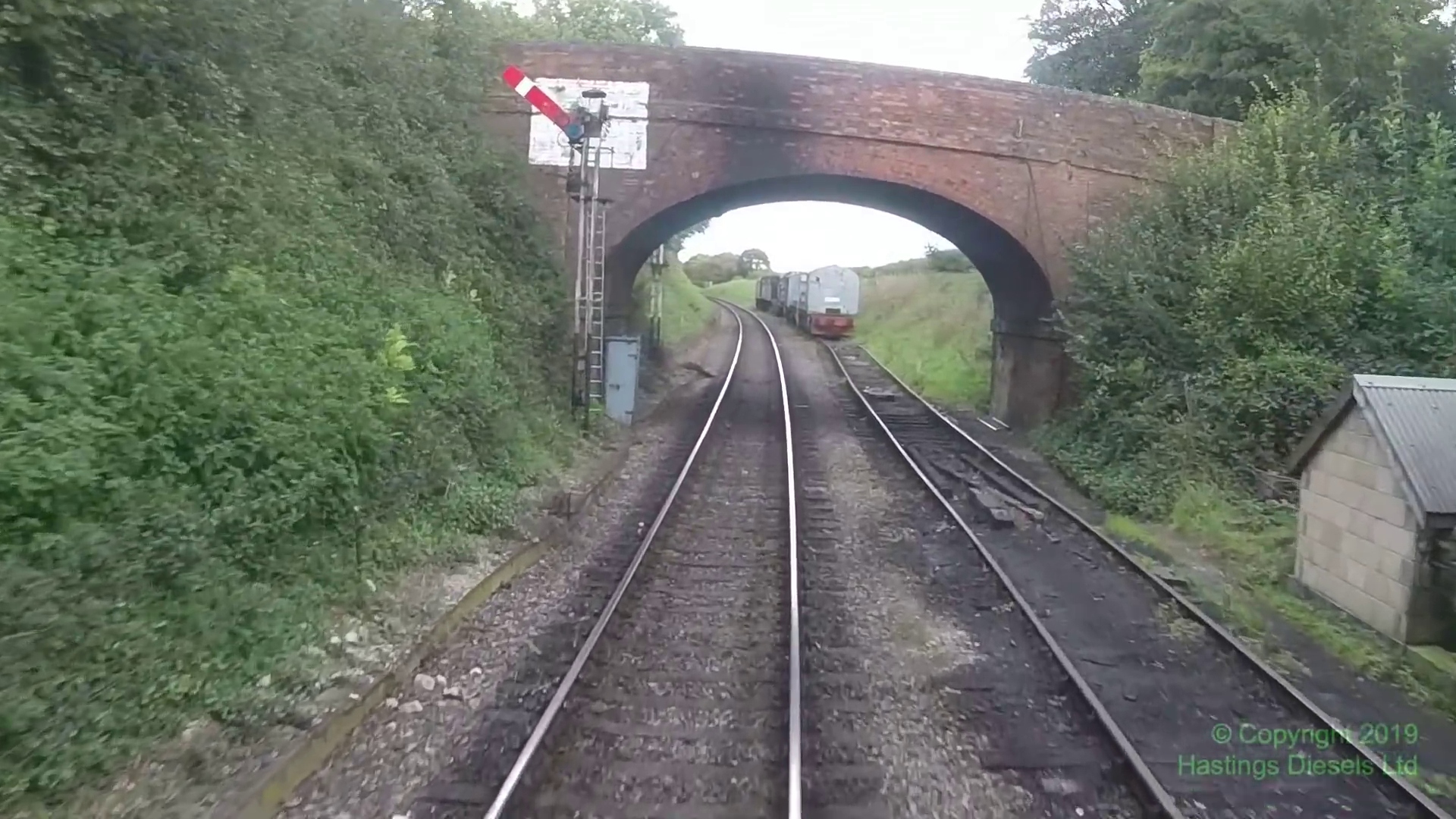}
		\label{pic1}} 
	\hfill
	\subfloat[Ground truth]{\includegraphics[width=2in]{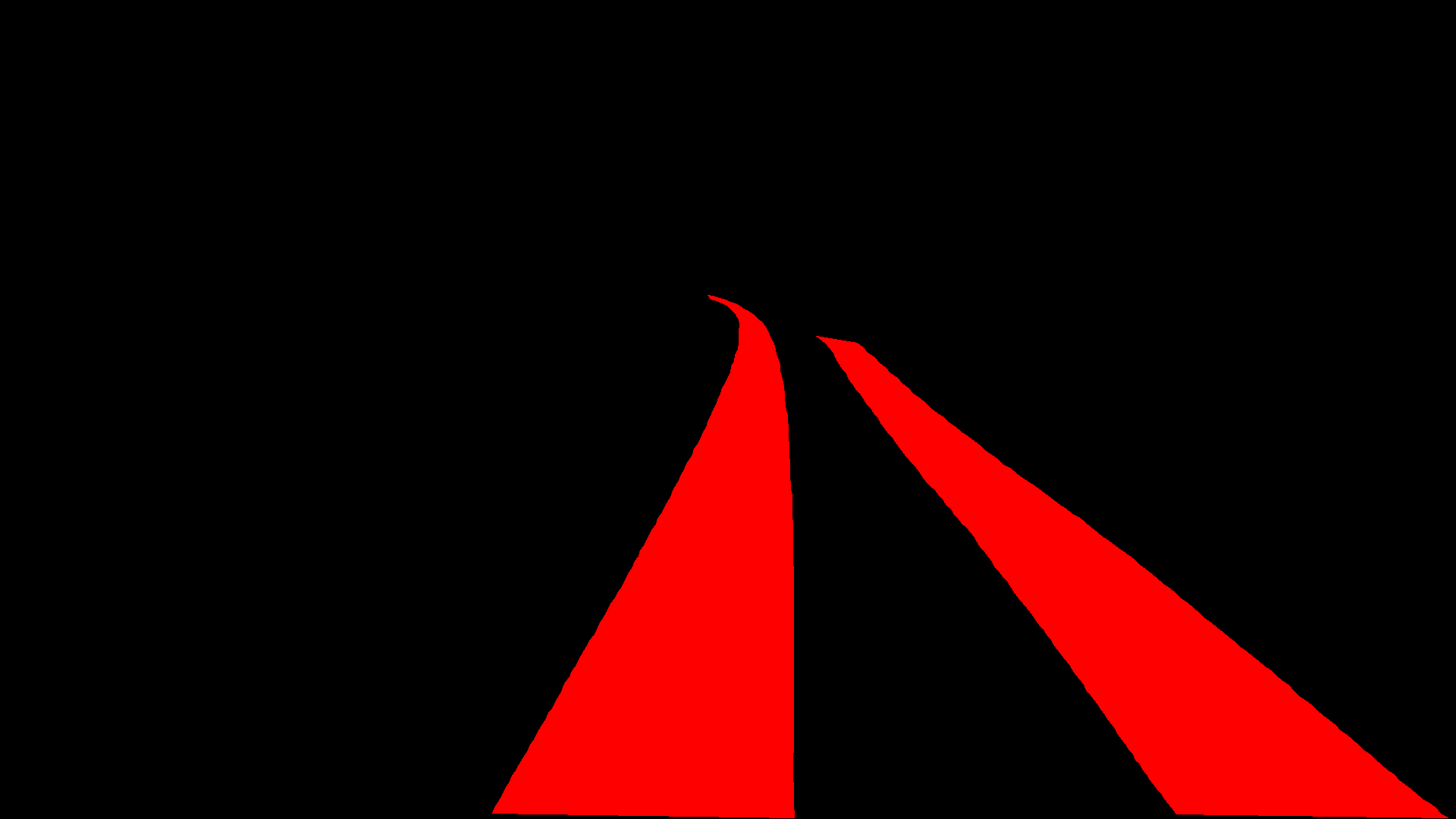}
		\label{pic2}}
	\hfill
	\subfloat[640×640 (baseline)]{\includegraphics[width=2in]{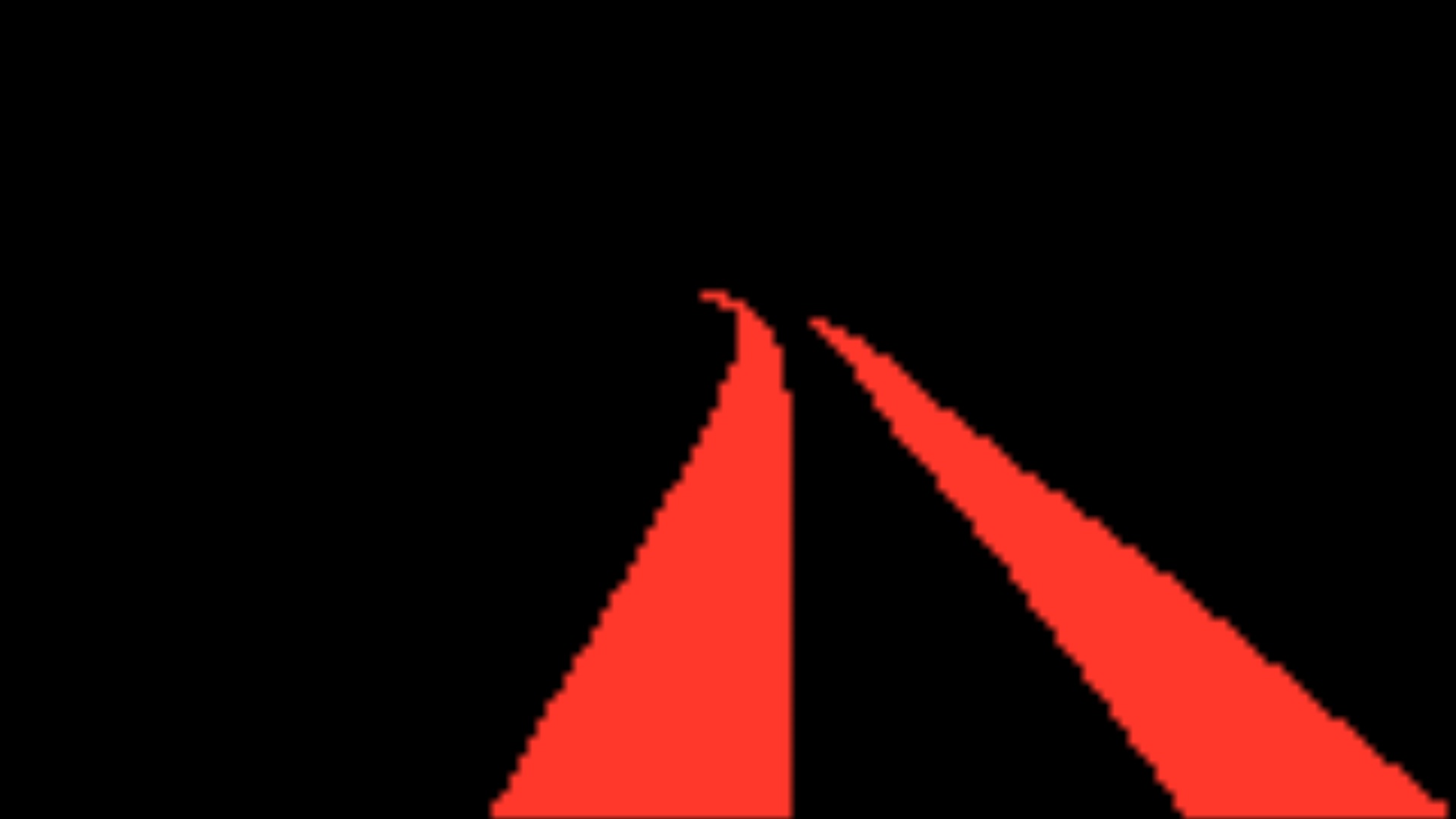}
		\label{pic3}}
	\hfill
	\subfloat[640×640 (FP16)]{\includegraphics[width=2in]{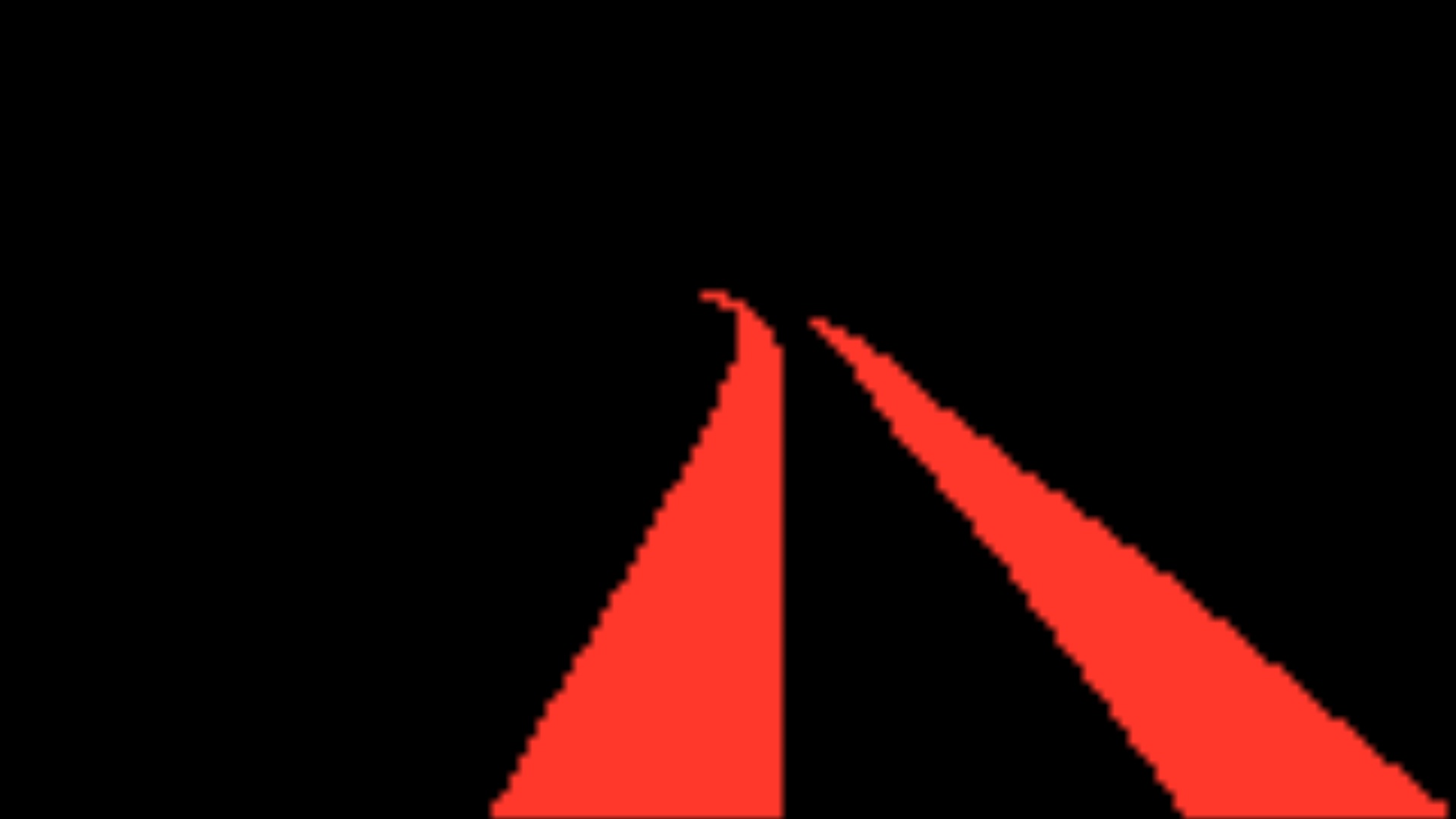}
		\label{pic4}}
	\hfill
	\subfloat[480×480 (FP16)]{\includegraphics[width=2in]{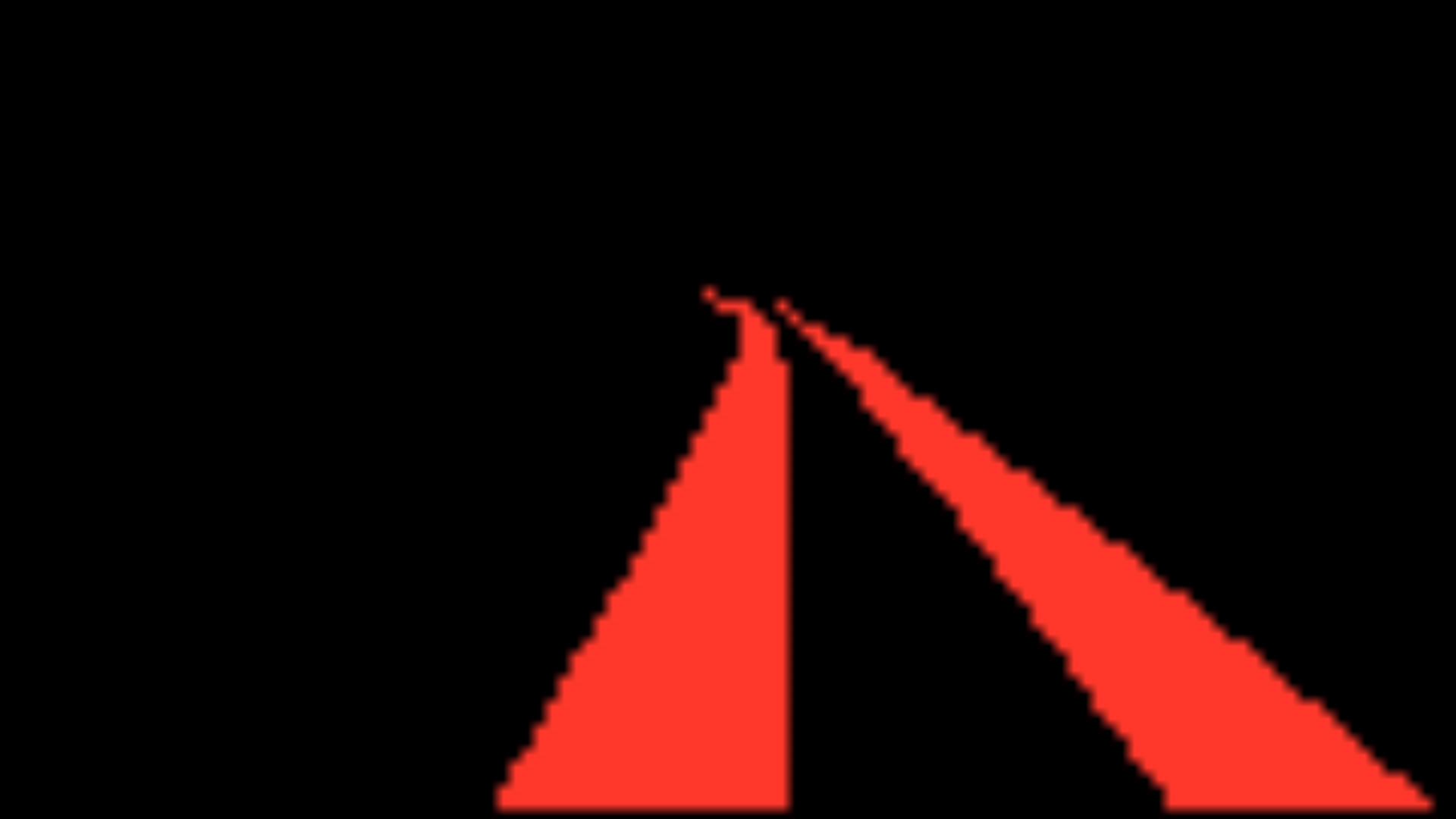}
		\label{pic5}}
	\hfill
	\subfloat[224×224 (FP16)]{\includegraphics[width=2in]{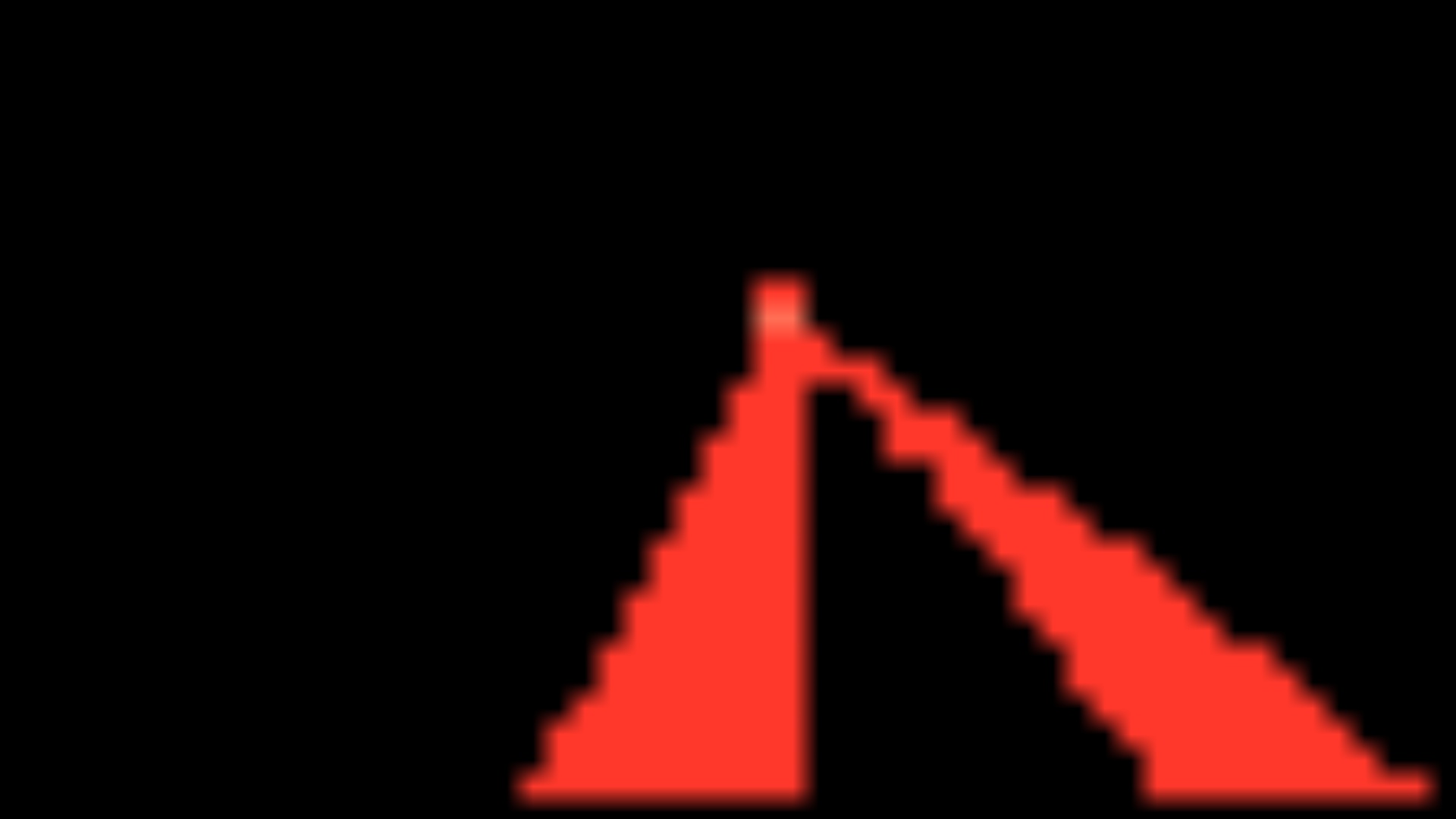}
		\label{pic6}}
	\hfill
	\caption{Visualization of experimental results under different input sizes on RailSem19.}
	\label{res_fig}
\end{figure*}

Fig. \ref*{res_fig} presents a comparison of the prediction results obtained from the proposed algorithm at various input sizes and precision. The first row displays, from left to right, the original image, the ground truth, and the image predicted with FP32-precision weights. The second row, also from left to right, displays the prediction results of the quantized model with FP16-precision when the static input size is 640×640, 480×480, and 224×224 respectively. It can be observed that when the input size is reduced to 480×480 and the quantization precision is FP16, the prediction results are not significantly different from those obtained with FP32 precision. However, when the input size is further reduced to 224×224, the segmentation at the track edge becomes less accurate, resulting in a jagged appearance.
\begin{table*}[!t]
	\caption{Ablation on detailed components in our approach on railsem19}\label{tab:a1} 
	\centering
	\begin{tabular}{ccc|cccc} 
		\toprule
		Efficient Head & C3Ghost & 2 Pred. & $mAP^{Mask}_{50}$	& GFLOPs & Params & Model Size (MB) \\ 
		\hline
		&	         &	          &0.847 &12.8	&3.41M	&6.8 \\
		\checkmark &         	&            &0.847 &10.5	&3.01M	&6.2 \\
		& \checkmark &            &0.838  &10.2	&2.66M	&5.6 \\
		\checkmark & \checkmark	&		     &0.838	 &8.8	&2.41M	&5.1 \\
		\checkmark & \checkmark	& \checkmark &0.833	 &8.0	&1.39M	&3.0 \\
		\toprule
	\end{tabular}
\end{table*}

\begin{table*}[!t]
	\caption{Ablation on FPN output layers on Railsem19}\label{tab:a2} 
	\centering
	\begin{tabular}{ccccc} 
		\toprule
		Method & $mAP^{Mask}_{50}$	& GFLOPs & Params  & Model Size (MB) \\ 
		\hline
		With P3	            &0.771	&6.9	&1.05M	&2.3 \\
		With P4, P5	        &0.736	&5.6	&2.58M	&5.4 \\
		With P3,P4,P5	    &0.838	&8.8	&2.42M	&5.1 \\
		With P3,P4 (Ours)	&0.833	&8.0	&1.39M	&3.0 \\
		\toprule
	\end{tabular}
\end{table*}

\begin{table*}[!t]
	\caption{Ablation on acceleration tricks}\label{tab:a3} 
	\centering
	\begin{tabular}{cccccccc} 
		\toprule
		Optimized model & FP16 & CUDA* & Pre- & infer & Post- & Total & FPS   \\
		\hline
		&      &       & 0.71 & 3.52  & 1.41  & 5.64  & 177.3 \\
		\checkmark               &      &       & 0.66 & 3.01  & 1.41  & 5.08  & 196.9 \\
		\checkmark               & \checkmark    &       & 4.50 & 1.30  & 1.67  & 7.47  & 133.9 \\
		\checkmark               & \checkmark    & \checkmark     & 0.45 & 0.97  & 0.10  & 1.52  & 657.9 \\
		\toprule
	\end{tabular}
\end{table*}

\subsection{Ablation Study}
To further analyze the effects of our proposed method, we conducted following comprehensive ablation experiments from the aspects of network design and model acceleration tricks.

\textbf{The impact of designed model components.} As illustrated in Table \ref*{tab:a1}, the ablation study primarily involves three variables: Efficient Head, C3Ghost, and 2 pred. These respectively indicate whether the Efficient Head is used to simplify the head structure of the original network, whether C3Ghost is utilized to construct the encode, and whether only two layers of features are used for prediction. when employing the Efficient Head, the model turns to require much less GFLOPs and params without sacrificing any accuracy. When refactoring the encoder using the C3Ghost module, the model accuracy decreased slightly by 0.009, while the model size and GFLOPs drop significantly by 17.65\% and 20.31\%, respectively. By incorporating Efficient Head on top of utilizing C3Ghost, it can be noted that the model accuracy remains unaffected while further reducing computational complexity, thereby highlighting the efficacy of Efficient Head in striking a balance between accuracy and speed. Subsequently, solely employing P3 and P4 for prediction on this basis results in a rapid reduction in both model size and GFLOPs, along with a marginal decline in accuracy, which is deemed acceptable for reduced calculations.

\textbf{The impact of FPN output layers.} In terms of scenes with narrow range of scale variations, multi-scale prediction does not improve the accuracy but slow down the inference. To delve into the impact of FPN multi-scale feature layer on model prediction, we carried out some experiments, and the results are presented in Table  \ref*{tab:a2}. It is clear that P3 takes on a dominant role. Abandoning P5 does not affect much in mAP,  but it leads to approximately 40\% and 10\% decreases in model size and GFLOPs respectively.

\textbf{The impact of acceleration tricks.} To further verify the effectiveness of different model acceleration tricks, we set the model input size to 640×640 and conducted ablation experiments on an RTX3060 Ti, as illustrated in Table \ref*{tab:a3}. We use three variables: 'Optimized model' indicates whether an optimized model is used, 'FP16' denotes the use of TensorRT for model quantization, and 'CUDA*' implies the use of CUDA for acceleration of pre- and post-processing. 'Pre-', 'Infer', 'Post-', and 'Total' represent the time spent on data pre-processing, model inference, data post-processing, and overall execution, respectively. Furthermore, to obtain more reliable results, the experiments were calculated 1,000 times and then averaged. The results indicate that the inference speed of the optimized model has been slightly improved, while the pre- and post-processing time remains largely unchanged. Building upon this, after quantizing the model to FP16 precision, the model inference speed significantly improves, dropping from 3.01 to 1.3 ms. However, the time consumption of the data pre-processing part increases from 0.66 to 4.5 ms. This is due to that TensorRT is built on CUDA, with data pre-processing and post-processing carried out on the CPU. When both CPU and GPU are required simultaneously in the processing flow, additional space application and data copying in memory and video memory are necessitated. When high computing performance devices are used for minor computations, data access and memory copying become the speed bottlenecks. To address this issue, we shifted pre- and post-processing to the GPU for computation. It is evident that the computation speeds of pre- and post-processing greatly improve after GPU acceleration, reducing the total execution time to only 1.5 ms, up to 657.9 FPS.

\section{Conclusion}
In this paper, we proposed an efficient network algorithm designed for railway scenarios, capable of accomplishing real-time track segmentation on low-performance edge devices. We incorporated Ghost convolution and constructed a lightweight decoupled head to simplify the network. Post-training quantization was implemented to enhance inference speed, and the pre- and post-processing of the model were further expedited through the GPU parallelism strategy. Our proposed method was validated on a publicly accessible railway datasets. Our research primarily concentrates on acceleration at the visual image and algorithmic levels. Future research might consider integrating multi-modal data to further improve model accuracy and explore acceleration at the hardware level.

\bibliographystyle{IEEEtran}
\bibliography{IEEEabrv,ref}

\end{document}